\documentclass[final]{cvpr}

\usepackage{times}
\usepackage{epsfig}
\usepackage{graphicx}
\usepackage{amsmath}
\usepackage{amssymb}
\usepackage{tabularx}
\usepackage{tabularx,booktabs}
\usepackage{appendix}
\newcolumntype{C}{>{\centering\arraybackslash}X} 

\usepackage{xcolor}
\definecolor{light-gray}{gray}{0.95}

\usepackage[pagebackref=true,breaklinks=true,colorlinks,bookmarks=false]{hyperref}

\usepackage{cuted}
\usepackage{capt-of}
\usepackage{subcaption}
\usepackage{adjustbox}
\begin{document}

\title{AdaBins: Depth Estimation using Adaptive Bins}

\author{Shariq Farooq Bhat \\
KAUST\\
{\tt\small shariq.bhat@kaust.edu.sa}

\and
Ibraheem Alhashim\\
KAUST\\
{\tt\small ibraheem.alhashim@kaust.edu.sa}

\and
Peter Wonka\\
KAUST\\
{\tt\small pwonka@gmail.com}
}

\maketitle

\begin{abstract}
We address the problem of estimating a high quality dense depth map from a single RGB input image. We start out with a baseline encoder-decoder convolutional neural network architecture and pose the question of how the global processing of information can help improve overall depth estimation. To this end, we propose a transformer-based architecture block that divides the depth range into bins whose center value is estimated adaptively per image. The final depth values are estimated as linear combinations of the bin centers. We call our new building block \emph{AdaBins}. Our results show a decisive improvement over the state-of-the-art on several popular depth datasets across all metrics. We also validate the effectiveness of the proposed block with an ablation study and provide the code and corresponding pre-trained weights of the new state-of-the-art model~\footnote{\url{https://github.com/shariqfarooq123/AdaBins}}.
\end{abstract}


\begin{figure}[t]
\centering
    \includegraphics[width=\linewidth]{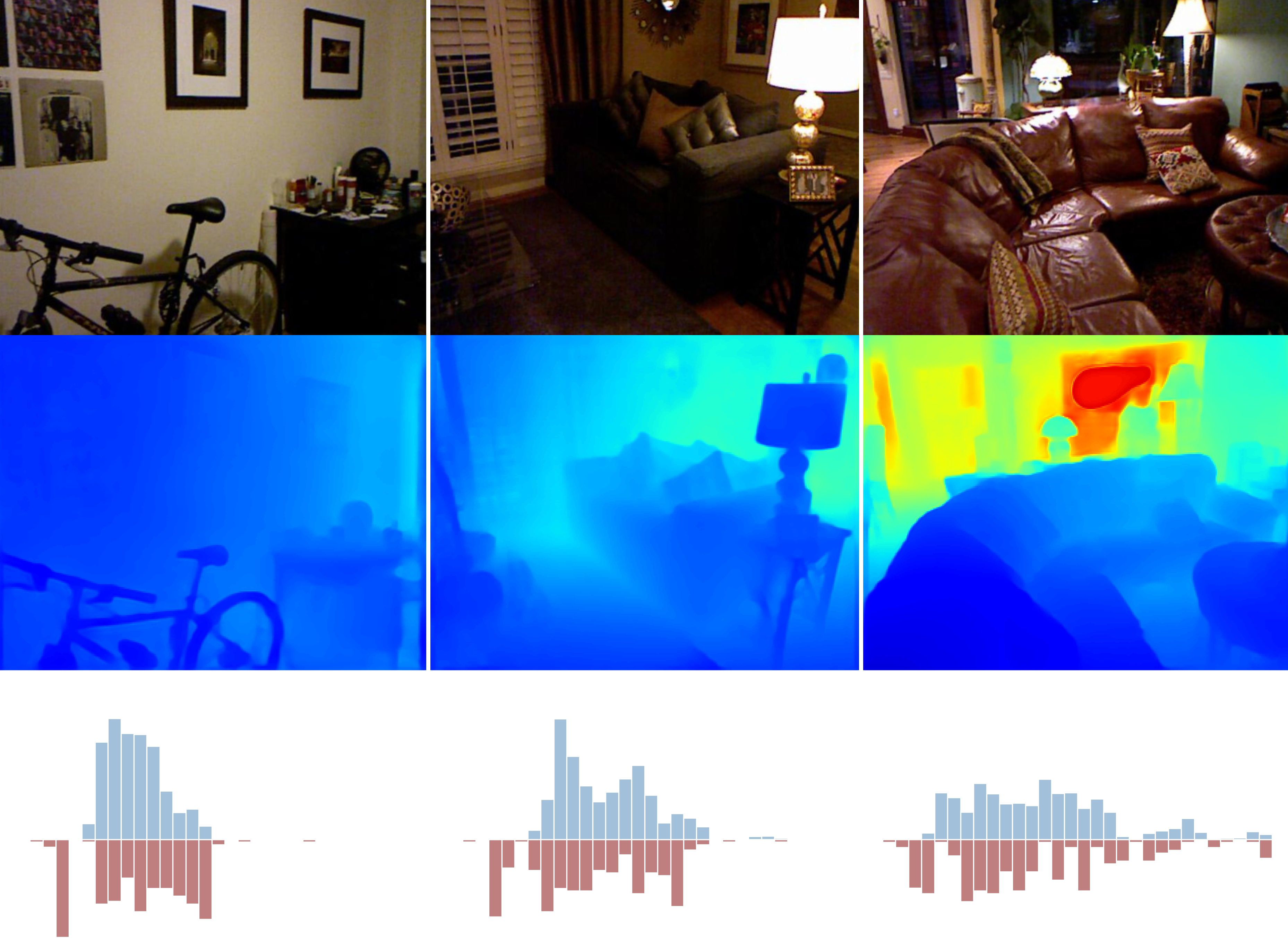}
    \captionof{figure}{Illustration of AdaBins: \textbf{Top}: input RGB images. \textbf{Middle}: depth predicted by our model. \textbf{Bottom}: histogram of depth values of the ground truth (blue) and histogram of the predicted adaptive depth-bin-centers (red) with depth values increasing from left to right. Note that the predicted bin-centers are focused near smaller depth values for closeup images but are widely distributed for images with a wider range of depth values.
    \label{fig:idea-illustration}}
\end{figure}

\begin{figure*}[t]
\centering
   \includegraphics[width=\linewidth]{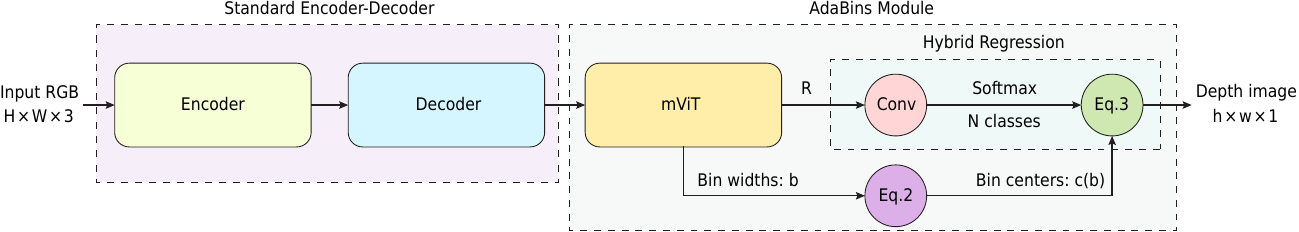}
   \caption{Overview of our proposed network architecture. Our architecture consists of two major components: an encoder-decoder block and our proposed adaptive bin-width estimator block called AdaBins. The input to our network is an RGB image of spatial dimensions $H$ and $W$, and the output is a single channel $h \times w$ depth image (e.g., half the spatial resolution).}
\label{fig:arch}
\end{figure*}

\section{Introduction}

This paper tackles the problem of estimating a high quality dense depth map from a single RGB input image. This is a classical problem in computer vision that is essential for many applications~\cite{Lee2011,moreno2007active,Hazirbas2016FuseNetID,DepthLab2020}.
In this work, we propose a new architecture building block, called \emph{AdaBins} that leads to a new state-of-the-art architecture for depth estimation on the two most popular indoor and outdoor datasets, NYU~\cite{Silberman2012} and KITTI~\cite{geiger2013vision}.

The motivation for our work is the conjecture that current architectures do not perform enough global analysis of the output values. A drawback of convolutional layers is that they only process global information once the tensors reach a very low spatial resolution at or near the bottleneck. However, we believe that global processing is a lot more powerful when done at high resolution. Our general idea is to perform a global statistical analysis of the output of a traditional encoder-decoder architecture and to refine the output with a learned post-processing building block that operates at the highest resolution. As a particular realization of this idea, we propose to analyze and modify the distribution of the depth values.

Depth distribution corresponding to different RGB inputs can vary to a large extent (see Fig.~\ref{fig:idea-illustration}). Some images have most of the objects located over a very small range of depth values. Closeup images of furniture will, for example, contain pixels most of which are close to the camera while other images may have depth values distributed over a much broader range, e.g. a corridor, where depth values range from a small value to the maximum depth supported by the network. Along with the ill-posed nature of the problem, such a variation in depth distribution makes depth regression in an end-to-end manner an even more difficult task. Recent works have proposed to exploit assumptions about indoor environments such as planarity constraints \cite{bts_lee2019big,dav_huynh2020guiding} to guide the network, which may or may not hold for a real-world environment, especially for outdoors scenes. 

Instead of imposing such assumptions, we investigate an approach where the network learns to adaptively \emph{focus} on regions of the depth range which are more probable to occur in the scene of the input image.

Our main contributions are the following:
\begin{itemize}
    \item We propose an architecture building block that performs global processing of the scene's information. We propose to divide the predicted depth range into bins where the bin widths change per image. The final depth estimation is a linear combination of the bin center values. 
    \item We show a decisive improvement for supervised single image depth estimation across all metrics for the two most popular datasets, NYU~\cite{Silberman2012} and KITTI~\cite{geiger2013vision}.
    \item We analyze our findings and investigate different modifications on the proposed AdaBins block and study their effect on the accuracy of the depth estimation.
\end{itemize}

\section{Related Work}
The problem of 3D scene reconstruction from RGB images is an ill-posed problem. Issues such as lack of scene coverage, scale ambiguities, translucent or reflective materials all contribute to ambiguous cases where geometry cannot be derived from appearance. Recently, methods that rely on convolutional neural networks (CNNs) are able to produce reasonable depth maps from a single RGB input image at real-time speeds.

\textbf{Monocular depth estimation} has been considered by many CNN methods as a regression of a dense depth map from a single RGB image \cite{Eigen2014,Laina2016,Xu2017,Hao2018DetailPD,Xu2018StructuredAG,Fu2018DeepOR,Hu2018RevisitingSI,Alhashim2018,bts_lee2019big,dav_huynh2020guiding}. 

As the two most important competitors, we consider BTS~\cite{bts_lee2019big}
and DAV~\cite{dav_huynh2020guiding}. BTS uses local planar guidance layers to guide the features to full resolution instead of standard upsampling layers during the decoding phase. DAV uses a standard encoder-decoder scheme and proposes to exploit co-planarity of objects in the scene via attention at the bottleneck. Our results section compares to these (and many other) methods.

\textbf{Encoder-decoder} networks have made significant contributions in many vision related problems such as image segmentation \cite{Ronneberger2015u}, optical flow estimation \cite{Dosovitskiy2015}, and image restoration \cite{LehtinenMHLKAA18}. In recent years, the use of such architectures have shown great success both in the supervised and the unsupervised setting of the depth estimation problem \cite{Godard2017,Ummenhofer2017,Huang2018DeepMVSLM,Zhou2018DeepTAMDT,Alhashim2018}. Such methods typically use one or more encoder-decoder networks as a sub part of their larger network. In this paper we adapted the baseline encoder-decoder network architecture used by \cite{Alhashim2018}. This allows us to more explicitly study the performance attribution of our proposed extension on the pipeline which is typically a difficult task.

\textbf{Transformer} networks are gaining greater attention as a viable building block outside of their traditional use in NLP tasks and into computer vision tasks \cite{pmlr-v80-parmar18a,Wang_2018_CVPR,detr2020,dosovitskiy2020}. Following the success of recent trends that combine CNNs with Transformers \cite{detr2020}, we propose to leverage a Transformer encoder as a building block for non-local processing on the output of a CNN.

\begin{figure}[t]
\centering
   \includegraphics[width=0.7\linewidth]{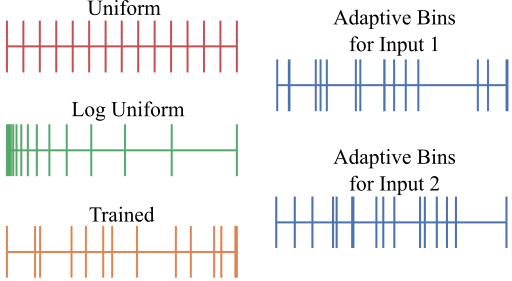}
   \caption{Choices for bin widths. Uniform and Log-uniform bins are pre-determined. `Trained bins' vary from one dataset to another. Adaptive bins vary for each input image.}
\label{fig:binchoices}
\end{figure}

\section{Methodology}

In this section, we present the motivation for this work, provide details of the AdaBins architecture, and describe the corresponding loss functions used.

\subsection{Motivation}
Our idea could be seen as a generalization of depth estimation via an ordinal regression network as proposed by Fu et al.~\cite{Fu2018DeepOR}. Fu et al. observed that a performance improvement could be achieved if the depth regression task is transformed into a classification task. They proposed to divide the depth range into a fixed number of bins of predetermined width. Our generalization solves multiple limitations of the initial approach.
First, we propose to compute adaptive bins that dynamically change depending on the features of the input scene. 
Second, a classification approach leads to a discretization of depth values which results in poor visual quality with obvious sharp depth discontinuities. This might still lead to good results with regard to the standard evaluation metrics, but it can present a challenge for downstream applications, e.g. computational photography or 3D reconstruction. Therefore, we propose to predict the final depth values as a linear combination of bin centers. This allows us to combine the advantages of classification with the advantages of depth-map regression.
Finally, compared to other architectures, e.g. DAV~\cite{dav_huynh2020guiding}, we compute information globally at a high resolution and not primarily in the bottleneck part at a low resolution.

\subsection{AdaBins design}
\label{sec:adabins-design-choices}

Here, we discuss four design choices of our proposed architecture that are most important for the obtained results.

First, we employ an adaptive binning strategy to discretize the depth interval $D = (d_{min}, d_{max})$ into $N$ bins. This interval is fixed for a given dataset and is determined by dataset specification or manually set to a reasonable range. To illustrate our idea of dividing a depth interval into bins, we would like to contrast our final solution with three other possible design choices we evaluated: 
\begin{itemize}
\item Fixed bins with a uniform bin width: the depth interval $D$ is divided into $N$ bins of equal size.
\item Fixed bins with a log scale bin width: the depth interval $D$ is divided into bins of equal size in log scale.
\item Trained bin widths: the bin widths are adaptive and can be learned for a particular dataset. While the bin widths are general, all images finally share the same bin subdivision of the depth interval $D$.
\item AdaBins: the bin widths $\textbf{b}$ are adaptively computed for each image.
\end{itemize}
We recommend the strategy of AdaBins as the best option and our ablation study validates this choice by showing the superiority of this design over its alternatives. An illustration of the four design choices for bin widths can be seen in Fig.~\ref{fig:binchoices}.

Second, discretizing the depth interval $D$ into bins and assigning each pixel to a single bin leads to depth discretization artifacts. We therefore predict the final depth as a linear combination of bin centers enabling the model to estimate smoothly varying depth values. 

Third, several previous architectures propose performing global processing using attention blocks to process information after an encoder block in the architecture (e.g., image captioning~\cite{Cornia_2020_CVPR, transf_obj_NEURIPS2019_680390c5} or object detection~\cite{detr2020}). Also, the current state-of-the-art in depth estimation uses this strategy~\cite{dav_huynh2020guiding}. Such an architecture consists of three blocks ordered as such: encoder, attention, followed by a decoder. We initially followed this approach but noticed that better results can be achieved when using attention at the spatially higher resolution tensors. We therefore propose an architecture that also has these three blocks, but ordered as follows: encoder, decoder, and finally attention.

Fourth, we would like to build on the simplest possible architecture to isolate the effects of our newly proposed \emph{AdaBins} concept. We therefore build on a modern encoder-decoder~\cite{Alhashim2018} using EfficientNet B5~\cite{efficentnet_TanL19} as the backbone for the encoder.

In the next subsection, we provide a description of the entire architecture.

\begin{figure}[t]
\centering
   \includegraphics[width=\linewidth]{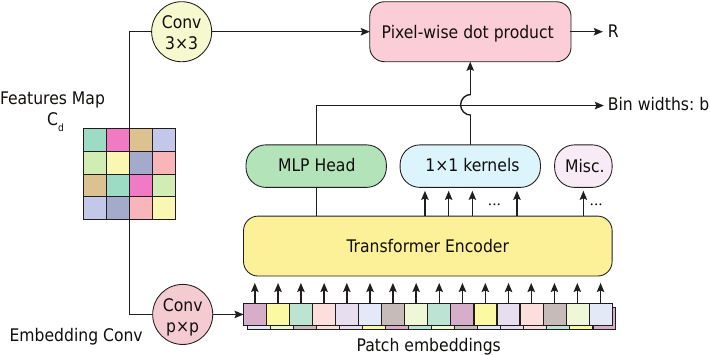}
   \caption{An overview of the mini-ViT block. The input to the block is a multi-channel feature map of the input image. The block includes a Transformer encoder that is applied on patch embeddings of the input for the purpose of learning to estimate bin widths $b$ and a set of convolutional kernels needed to compute our Range-Attention-Maps $R$.}
\label{fig:mVit-module}
\end{figure}

\subsection{Architecture description}
Fig.~\ref{fig:arch} shows an overview of our proposed depth estimating architecture. Our architecture consists of two major components: 1) an encoder-decoder block built on a pre-trained EfficientNet B5~\cite{efficentnet_TanL19} encoder and a standard feature upsampling decoder; 2) our proposed adaptive bin-width estimator block called AdaBins.
The first component is primarily based on the simple depth regression network of Alhashim and Wonka~\cite{Alhashim2018} with some modifications. The two basic modifications are switching the encoder from DenseNet~\cite{huang2017densely} to EfficientNet B5 and using a different appropriate loss function for the new architecture. In addition, the output of the decoder is a tensor $\textbf{x}_d \in \mathbb{R}^{h \times w \times C_d}$, not a single channel image representing the final depth values. We refer to this tensor as the ``\textit{decoded features}". 
The second component is a key contribution in this paper, the AdaBins module. The input to the AdaBins module are \textit{decoded features} of size $h \times w \times C_d$ and the output tensor is of size $h \times w \times 1$.
Due to memory limitations of current GPU hardware, we use $h=H/2$ and $w=W/2$ to facilitate better learning with larger batch sizes. The final depth map is computed by simply bilinearly upsampling to $H\times W\times 1$. 

The first block in the AdaBins module is called mini-ViT. An overview of this block is shown in Fig.~\ref{fig:mVit-module}. It is a simplified version of a recently proposed technique of using transformers for image recognition~\cite{dosovitskiy2020} with minor modifications. The details of mini-ViT are explained in the next paragraph. There are two outputs of mini-ViT: 1) a vector $\textbf{b}$ of bin-widths, which defines how the depth interval $D$ is to be divided for the input image, and 2) Range-Attention-Maps $\mathcal{R}$ of size $h \times w \times C$, that contain useful information for pixel-level depth computation.

\begin{table}[t]
\centering
\begin{tabular}{@{}lllllll@{}}
\toprule
\begin{tabular}[c]{@{}l@{}}Patch \\ size ($p$)\end{tabular} & E   & Layers & \begin{tabular}[c]{@{}l@{}}num\\ heads\end{tabular} & C   & \begin{tabular}[c]{@{}l@{}}MLP \\ Size\end{tabular} & Params \\ \midrule
16    & 128 & 4      & 4    & 128 & 1024      & 5.8 M \\ \bottomrule
\end{tabular}
\caption{Mini-ViT architecture details.}
\label{tab:arch-mvit}
\end{table}

\paragraph{Mini-ViT.} Estimating sub-intervals within the depth range $D$ which are more probable to occur for a given image would require a combination of local structural information and global distributional information at the same time. We propose to use global attention in order to calculate a bin-widths vector $\textbf{b}$ for each input image. Global attention is expensive both in terms of memory and computational complexity, especially at higher resolutions. However, recent rapid advances in transformers provide some efficient alternatives. We take inspiration from the Vision Transformer ViT~\cite{dosovitskiy2020} in designing our AdaBins module with transformers. We also use a much smaller version of the transformer proposed as our dataset is smaller and refer to this transformer as mini-ViT or mViT in the following description.

\paragraph{Bin-widths.} We first describe how the bin-widths vector $\textbf{b}$ is obtained using mViT. The input to the mViT block is a tensor of \textit{decoded features}~$\mathbf{x_d}\in~\mathbb{R}^{h \times w \times C_d}$. However, a transformer takes a sequence of fixed size vectors as input. We first pass the \textit{decoded features} through a convolutional block, named as \textit{Embedding Conv} (see Fig~\ref{fig:mVit-module}), with kernel size $p\times p$, stride $p$ and number of output channels $E$. Thus, the result of this convolution is a tensor of size $h/p~\times~w/p~\times E$ (assuming both $h$ and $w$ are divisible by $p$). The result is reshaped into a spatially flattened tensor $\mathbf{x_p}~\in~\mathbb{R}^{S\times E}$, where $S=\frac{hw}{p^2}$ serves as the effective sequence length for the transformer. We refer to this sequence of $E$-dimensional vectors as \textit{patch embeddings}.

Following common practice~\cite{detr2020,dosovitskiy2020}, we add learned positional encodings to the patch embeddings before feeding them to the transformer. Our transformer is a small transformer encoder (see Table.~\ref{tab:arch-mvit} for details) and outputs a sequence of \textit{output embeddings} $\mathbf{x_o}~\in~\mathbb{R}^{S\times E}$. We use an MLP head over the first output embedding (we also experimented with a version that has an additional special token as first input, but did not see an improvement). The MLP head uses a ReLU activation and outputs an N-dimensional vector $\textbf{b}'$. Finally, we normalize the vector $\textbf{b}'$ such that it sums up to $1$, to obtain the bin-widths vector $\textbf{b}$ as follows:
\begin{equation}
    b_i = \frac{b'_i + \epsilon}{\sum_{j=1}^{N} (b'_j + \epsilon)},
\end{equation}
where $\epsilon=10^{-3}$. The small positive $\epsilon$ ensures each bin-width is strictly positive. The normalization introduces a competition among the bin-widths and conceptually forces the network to \textit{focus} on sub-intervals within $D$ by predicting smaller bin-widths at interesting regions of $D$. 

In the next subsection, we describe how the Range-Attention-Maps $\mathcal{R}$ are obtained from the \textit{decoded features} and the transformer output embeddings.

\paragraph{Range attention maps.} At this point, the \textit{decoded features} represent a high-resolution and local pixel-level information while the transformer output embeddings effectively contain more global information. As shown in Fig.~\ref{fig:mVit-module}, output embeddings 2 through $C+1$ from the transformer are used as a set of $1\times 1$ convolutional kernels and are convolved with the \textit{decoded features} (following a $3\times 3$ convolutional layer) to obtain the Range-Attention Maps $\mathcal{R}$. This is equivalent to calculating the Dot-Product attention weights between pixel-wise features treated as `keys' and transformer output embeddings as `queries'. This simple design of using output embeddings as convolutional kernels lets the network integrate adaptive global information from the transformer into the local information of the \textit{decoded features}. $\mathcal{R}$ and $\textbf{b}$ are used together to obtain the final depth map.

\paragraph{Hybrid regression.} Range-Attention Maps $\mathcal{R}$ are passed through a $1\times 1$ convolutional layer to obtain $N$-channels which is followed by a Softmax activation. We interpret the $N$ Softmax scores $p_k$, $k=1,...,N$, at each pixel as probabilities over $N$ depth-bin-centers $c(\textbf{b}) := \{c(b_1), c(b_2), ..., c(b_N)\}$
calculated from bin-widths vector $\textbf{b}$ as follows:
\begin{equation}
    c(b_i) =  d_{min} + (d_{max} - d_{min})(b_i/2 + \sum_{j=1}^{i-1} b_j)
\end{equation}

\begin{figure}
    \centering
    \begin{tabular}{l}
         \includegraphics[width=0.9\linewidth]{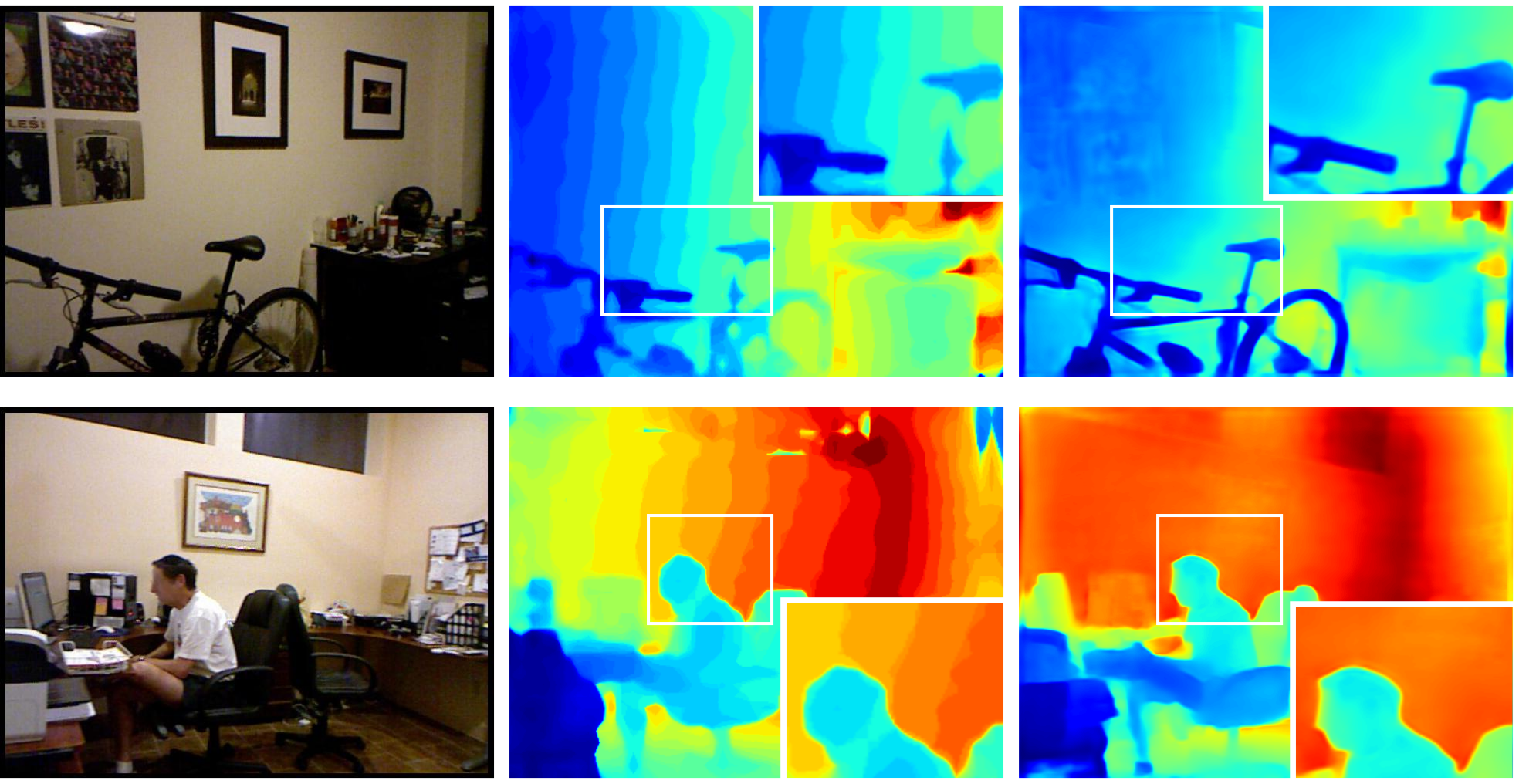} \\
        \hspace{0.1\linewidth}RGB \hspace{0.15\linewidth}Fu~\etal~\cite{Fu2018DeepOR}\hspace{0.15\linewidth}Ours\\ 
    \end{tabular}
    
    \caption{Demonstration of artifacts introduced by the discretization of the depth interval. Our hybrid regression results in smoother depth maps.}
    \label{fig:compare-DORN}
\end{figure}

Finally, at each pixel, the final depth value $\Tilde{d}$ is calculated from the linear combination of Softmax scores at that pixel and the depth-bin-centers $c(\textbf{b})$ as follows:
\begin{equation}
    \Tilde{d} = \sum_{k=1}^N c(b_k) p_k
\end{equation}
Compared to Fu et al.~\cite{Fu2018DeepOR} we do not predict the depth as the bin center of the most likely bin. This enables us to predict smooth depth maps without the discretization artifacts as can bee seen in Fig.~\ref{fig:compare-DORN}.

\subsection{Loss function}
\textbf{Pixel-wise depth loss.} Inspired by~\cite{bts_lee2019big}, we use a scaled version of the Scale-Invariant loss (SI) introduced by Eigen~\etal~\cite{Eigen2014}: 
\begin{equation}
    \mathcal{L}_{pixel} = \alpha \sqrt{\frac{1}{T}\sum_i g_{i}^{2} - \frac{\lambda}{T^2}(\sum_i g_i)^2}
    \label{eq:pixel-loss}
\end{equation}
where $g_i = \log \Tilde{d_i} - \log d_i$ and the ground truth depth $d_i$ and $T$ denotes the number of pixels having valid ground truth values. We use $\lambda = 0.85$ and $\alpha = 10$ for all our experiments.

\textbf{Bin-center density loss.} This loss term encourages the distribution of bin centers to follow the distribution of depth values in the ground truth. We would like to encourage the bin centers to be close to the actual ground truth depth values and the other way around. We denote the set of bin centers as $c(\textbf{b})$ and the set of all depth values in the ground truth image as $X$ and use the bi-directional Chamfer Loss ~\cite{Fan2017_chamfer_loss} as a regularizer:
\begin{equation}
    \mathcal{L}_{bins} = chamfer(X, c(\textbf{b})) + chamfer(c(\textbf{b}), X)
    \label{eq:chamfer-loss}
\end{equation}
Finally, we define the total loss as:
\begin{equation}
    \mathcal{L}_{total} = \mathcal{L}_{pixel} + \beta \mathcal{L}_{bins}
\end{equation}

We set $\beta = 0.1$ for all our experiments. We experimented with different loss functions including the RMSE loss, and the combined SSIM~\cite{Wang2004SSIM} plus $L_1$ loss suggested by~\cite{Alhashim2018}. However, we were able to achieve the best results with our proposed loss. We offer a comparison of the different loss functions and their performance in our ablation study.

\begin{table*}[t]
\centering
\begin{tabularx}{0.8\linewidth}{@{}l*{5}{C}c@{}}
\toprule
Method        & \textbf{$\delta_1$}$\uparrow$       & \textbf{$\delta_2$}$\uparrow$          & \textbf{$\delta_3$}~$\uparrow$            & REL~$\downarrow$          & RMS~$\downarrow$  & $log_{10}$~$\downarrow$ \\ \midrule
Eigen~\etal~\cite{Eigen2014}                                                       & 0.769          & 0.950          & 0.988          & 0.158            & 0.641          & --              \\ 
Laina~\etal~\cite{Laina2016}                                                   & 0.811          & 0.953          & 0.988          & 0.127            & 0.573          & 0.055                \\ 
Hao~\etal~\cite{Hao2018DetailPD}                                                      & 0.841          & 0.966          & 0.991          & 0.127            & 0.555          & 0.053                \\ 
Lee~\etal~\cite{Lee2011}                                                      & 0.837          & 0.971          & 0.994          & 0.131            & 0.538          &   --                   \\
Fu~\etal~\cite{Fu2018DeepOR}                                                     & 0.828          & 0.965          & 0.992          & 0.115            & 0.509          &   0.051                   \\
SharpNet~\cite{Ramamonjisoa_2019_ICCV}                                                            & 0.836          & 0.966          & 0.993          & 0.139            & 0.502          &      \underline{0.047}          \\
Hu~\etal~\cite{Hu2018RevisitingSI}                                                               & 0.866          & 0.975          & 0.993          & 0.115            & 0.530          &    0.050            \\ 
Chen~\etal~\cite{ijcai2019-98}                                                      & 0.878          & 0.977          & 0.994          & 0.111            & 0.514          &  0.048              \\ 
Yin~\etal~\cite{Yin_2019_ICCV}                                                              & 0.875          & 0.976          & 0.994          & \underline{0.108}            & 0.416          & 0.048               \\ 
BTS~\cite{bts_lee2019big}                                                              & \underline{0.885}          & 0.978          & 0.994          & 0.110            & \underline{0.392}          & \underline{0.047}          \\ 
DAV~\cite{dav_huynh2020guiding}                                                              & 0.882          & \underline{0.980}          & \underline{0.996} & \underline{0.108}            & 0.412          & --              \\ 
\midrule

\textbf{AdaBins (Ours)} & \textbf{0.903} & \textbf{0.984} & \textbf{0.997} & \textbf{0.103}     & \textbf{0.364} & \textbf{0.044} \\ 
\bottomrule
\end{tabularx}
\caption{Comparison of performances on the NYU-Depth-v2 dataset. The reported numbers are from the corresponding original papers. Best results are in bold, second best are underlined.}
\label{tab:results-nyu}
\end{table*}

\begin{table*}[t]
\centering
\begin{tabularx}{\textwidth}{@{}l*{6}{C}c@{}}
\toprule
Method            & \textbf{$\delta_1$}$\uparrow$    & \textbf{$\delta_2$}$\uparrow$    & \textbf{$\delta_3$}$\uparrow$    & REL~$\downarrow$ & Sq Rel~$\downarrow$ & RMS~$\downarrow$  & RMS log~$\downarrow$ \\ \midrule
Saxena~\etal~\cite{Saxena2005}     & 0.601 & 0.820  & 0.926 & 0.280   & 3.012  & 8.734 & 0.361    \\
Eigen~\etal~\cite{Eigen2014}      & 0.702 & 0.898 & 0.967 & 0.203   & 1.548  & 6.307 & 0.282    \\
Liu~\etal~\cite{Liu2016LearningDF}       & 0.680 & 0.898 & 0.967 & 0.201   & 1.584  & 6.471 & 0.273    \\
Godard~\etal~\cite{Godard2017}      & 0.861 & 0.949 & 0.976 & 0.114   & 0.898  & 4.935 & 0.206    \\
Kuznietsov~\etal~\cite{Kuznietsov2017} & 0.862 & 0.960  & 0.986 & 0.113   & 0.741  & 4.621 & 0.189    \\
Gan~\etal~\cite{Gan2018}        & 0.890 & 0.964 & 0.985 & 0.098   & 0.666  & 3.933 & 0.173    \\
Fu~\etal~\cite{Fu2018DeepOR}          & 0.932 & 0.984 & 0.994 & 0.072   & 0.307  & \underline{2.727} & 0.120    \\
Yin~\etal~\cite{Yin_2019_ICCV}        & 0.938 & 0.990  & 0.998 & 0.072   & --      & 3.258 & 0.117    \\
BTS\cite{bts_lee2019big}               & \underline{0.956} & \underline{0.993} & \underline{0.998} & \underline{0.059}   & \underline{0.245}  & 2.756 & \underline{0.096}    \\ 
\midrule
\textbf{AdaBins (Ours)} & \textbf{0.964} & \textbf{0.995} & \textbf{0.999} & \textbf{0.058}   & \textbf{0.190}  & \textbf{2.360} & \textbf{0.088} \\ \bottomrule
\end{tabularx}
\caption{Comparison of performances on the KITTI dataset. We compare our network against the state-of-the-art on this dataset. The reported numbers are from the corresponding original papers. Measurements are made for the depth range from $0m$ to $80m$. Best results are in bold, second best are underlined.}
\label{tab:results-kitti}
\end{table*}

\begin{table}[t]
\centering
\begin{adjustbox}{width=\linewidth,center}

\begin{tabular}{@{}lllllll@{}}
\toprule
\textbf{Loss}        & \textbf{$\delta_1$}$\uparrow$       & \textbf{$\delta_2$}$\uparrow$          & \textbf{$\delta_3$}$\uparrow$            & REL$\downarrow$          & RMS$\downarrow$  & $log_{10}$$\downarrow$ \\ \midrule
$L_1$/SSIM                                                        & 0.888          & 0.980          & 0.995          & 0.107            & 0.384          & 0.046          \\
SI                                                            & 0.897          & \textbf{0.984} & \textbf{0.997} & 0.106            & 0.368          & \textbf{0.044} \\
\begin{tabular}[c]{@{}l@{}}\textbf{SI+Bins}\end{tabular} & \textbf{0.903} & \textbf{0.984} & \textbf{0.997} & \textbf{0.103}   & \textbf{0.364} & \textbf{0.044} \\
\bottomrule
\end{tabular}
\end{adjustbox}
\caption{Comparison of performance with respect to the choice of loss function.}
\label{tab:ablation-loss}
\end{table}

\section{Experiments}
We conducted an extensive set of experiments on the standard depth estimation from a single image datasets for both indoor and outdoor scenes. In the following, we first briefly describe the datasets and the evaluation metrics, and then present quantitative comparisons to the state-of-the-art in supervised monocular depth estimation.

\subsection{Datasets and evaluation metrics}

\paragraph{NYU Depth v2} is a dataset that provides images and depth maps for different indoor scenes captured at a pixel resolution of $640\times480$ \cite{Silberman2012}. The dataset contains 120K training samples and 654 testing samples \cite{Eigen2014}. We train our network on a 50K subset. The depth maps have an upper bound of 10 meters.  Our network outputs depth prediction having a resolution of $320\times240$ which we then upsample by $2\times$ to match the ground truth resolution during both training and testing. We evaluate on the pre-defined center cropping by Eigen et al. \cite{Eigen2014}. At test time, we compute the final output by taking the average of an image's prediction and the prediction of its mirror image which is commonly used in previous work.

\paragraph{KITTI} is a dataset that provides stereo images and corresponding 3D laser scans of outdoor scenes captured using equipment mounted on a moving vehicle \cite{geiger2013vision}. The RGB images have a resolution of around $1241\times376$ while the corresponding depth maps are of very low density with lots of missing data. We train our network on a subset of around 26K images, from the left view, corresponding to scenes not included in the 697 test set specified by \cite{Eigen2014}. The depth maps have an upper bound of 80 meters. We train our network on a random crop of size $704\times352$. For evaluation, we use the crop as defined by Garg~\etal~\cite{garg10.1007/978-3-319-46484-8_45} and bilinearly upsample the prediction to match the ground truth resolution. The final output is computed by taking the average of an image's prediction and the prediction of its mirror image.

\paragraph{SUN RGB-D} is an indoor dataset consisting of around $10K$ images with high scene diversity collected with four different sensors~\cite{Song2015_sunrgbd,sun6751312,janoch2013category}. We use this dataset only for cross-evaluating pre-trained models on the official test set of 5050 images. We do not use it for training.

\paragraph{Evaluation metrics.} We use the standard six metrics used in prior work \cite{Eigen2014} to compare our method against state-of-the-art. These error metrics are defined as:
average relative error (REL): $\frac{1}{n}\sum_p^n \frac{\lvert y_p-\hat{y}_p \rvert}{y}$;
root mean squared error (RMS): $\sqrt{\frac{1}{n}\sum_p^n (y_p-\hat{y}_p)^2)}$;
average ($\log_{10}$) error: $\frac{1}{n}\sum_p^n \lvert \log_{10}(y_p)-\log_{10}(\hat{y}_p) \rvert$;
threshold accuracy ($\delta_i$): $\%$ of $y_p$ s.t. $\text{max}(\frac{y_p}{\hat{y}_p},\frac{\hat{y}_p}{y_p}) = \delta < thr$ for $thr=1.25,1.25^2,1.25^3$;
where $y_p$ is a pixel in depth image $y$, $\hat{y}_p$ is a pixel in the predicted depth image $\hat{y}$, and $n$ is the total number of pixels for each depth image. 
Additionally for KITTI, we use the two standard metrics: Squared Relative Difference (Sq.~Rel): $\frac{1}{n}\sum_p^n \frac{\|y_p-\hat{y}_p \|^2}{y}$; 
and RMSE log: $\sqrt{\frac{1}{n}\sum_p^n \|\log y_p - \log \hat{y}_p\|^2}$.

\subsection{Implementation details}
We implement the proposed network in PyTorch~\cite{NEURIPS2019_bdbca288}. For training, we use the AdamW optimizer~\cite{Loshchilov2019DecoupledWD} with weight-decay $10^{-2}$. We use the 1-cycle policy~\cite{DBLP:journals/corr/abs-1708-07120} for the learning rate with $max\_lr~=~3.5\times~10^{-4}$, linear warm-up from $max\_lr/25$ to $max\_lr$ for the first $30\%$ of iterations followed by cosine annealing to $max\_lr/75$. Total number of epochs is set to $25$ with batch size $16$. Training our model takes 20 min per epoch on a single node with four NVIDIA V100 32GB GPUs. For all results presented we train for 25 epochs. 
Our main model has about 78M parameters: 28M for the CNN encoder, 44M for the CNN decoder, and 5.8M for the new AdaBins module.

\begin{figure}[t]
    \centering
    \includegraphics[width=0.7\linewidth]{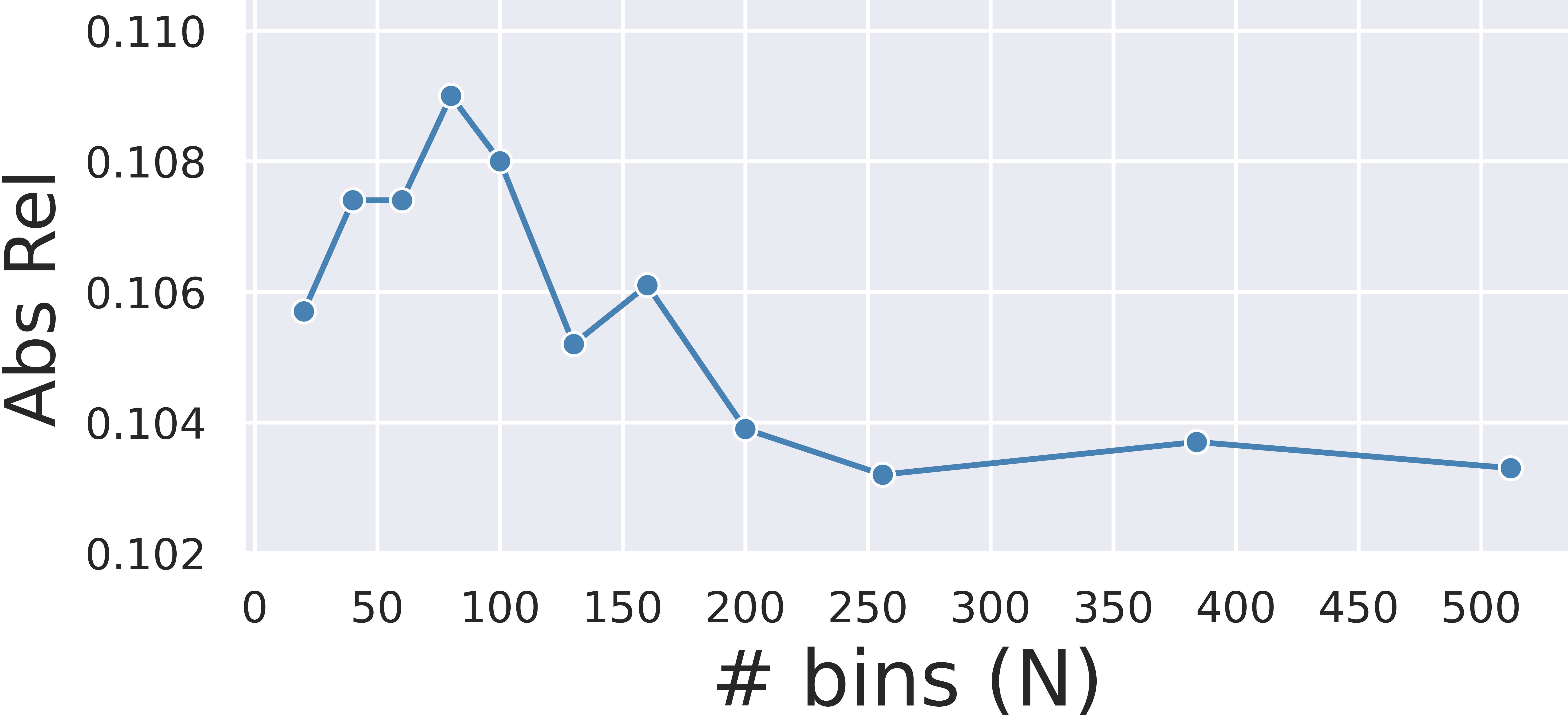}
    \caption{Effect of number of bins (N) on performance as measured by Absolute Relative Error metric. we can observe interesting behaviour for lower values of N. As N increases, performance starts to saturate.}
    \label{fig:ablation-N}
\end{figure}

\subsection{Comparison to the state-of-the-art}

\begin{figure*}
     \centering
     \subcaptionbox{RGB}{\includegraphics[height=7.5cm]{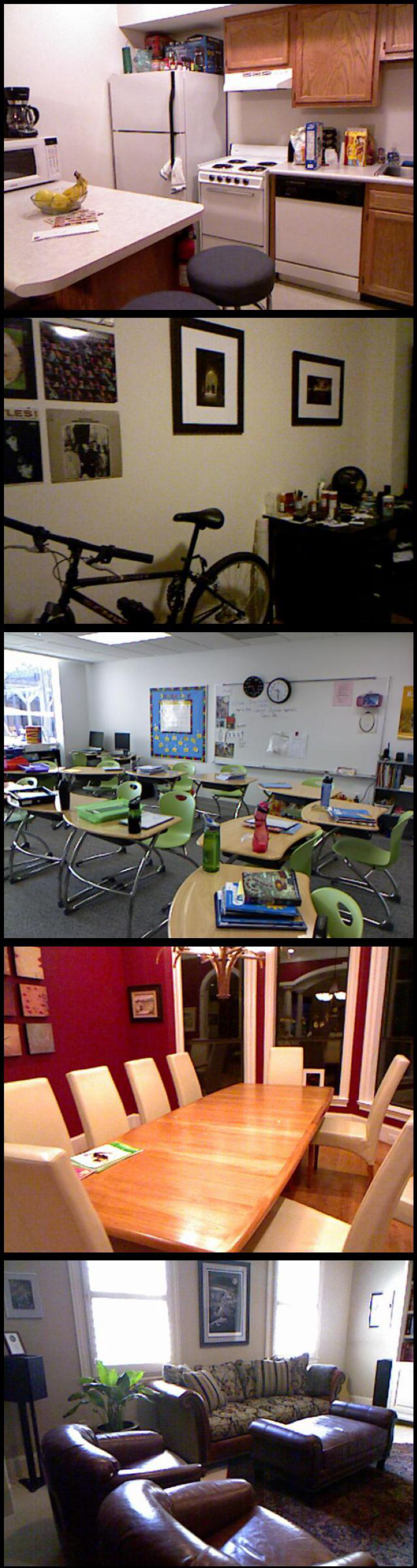}}\hspace{0.3em}%
     \subcaptionbox{BTS~\cite{bts_lee2019big}}{\includegraphics[height=7.5cm]{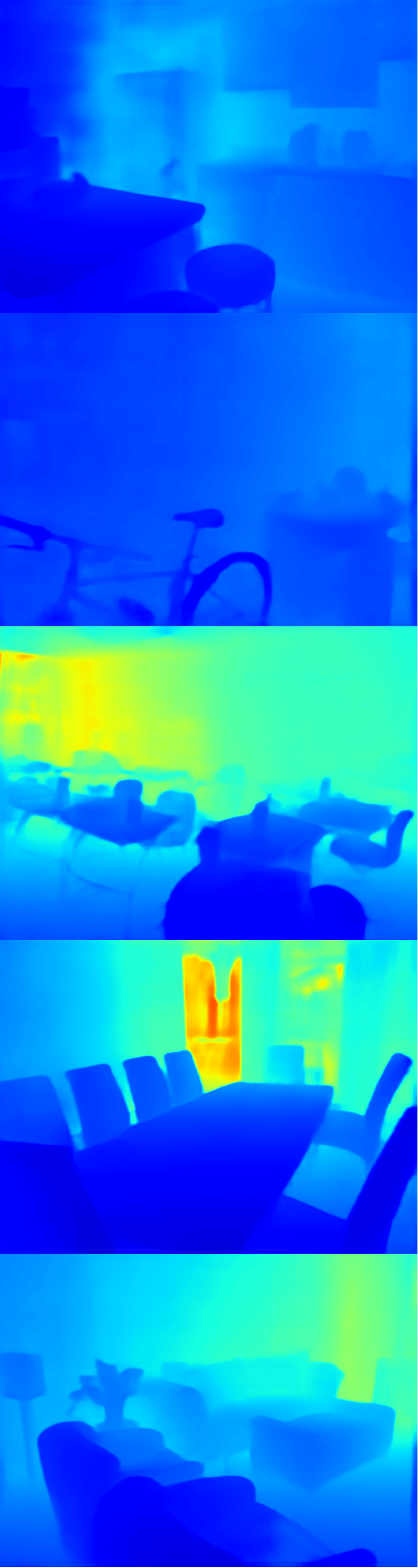}}\hspace{0.3em}%
     \subcaptionbox{DAV~\cite{ijcai2019-98}}{\includegraphics[height=7.5cm]{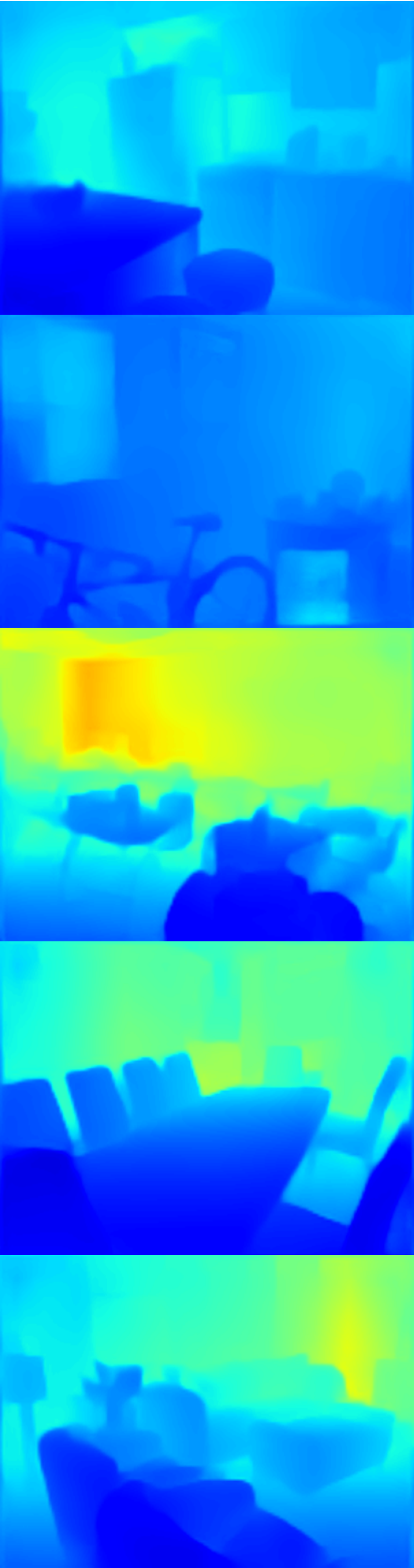}}\hspace{0.3em}%
     \subcaptionbox{Ours}{\includegraphics[height=7.5cm]{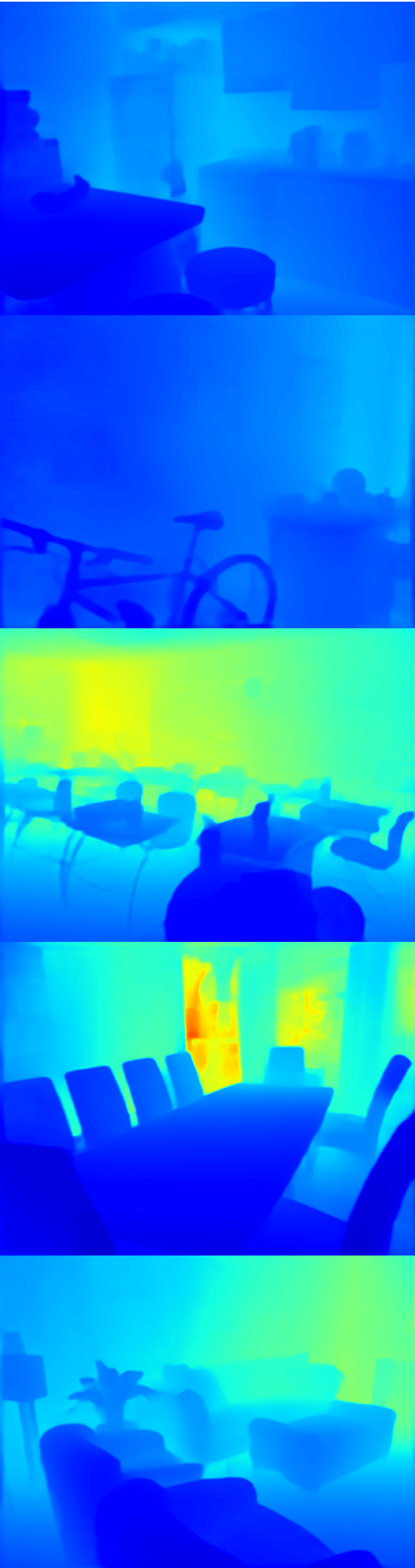}}\hspace{0.3em}%
     \subcaptionbox{GT}{\includegraphics[height=7.5cm]{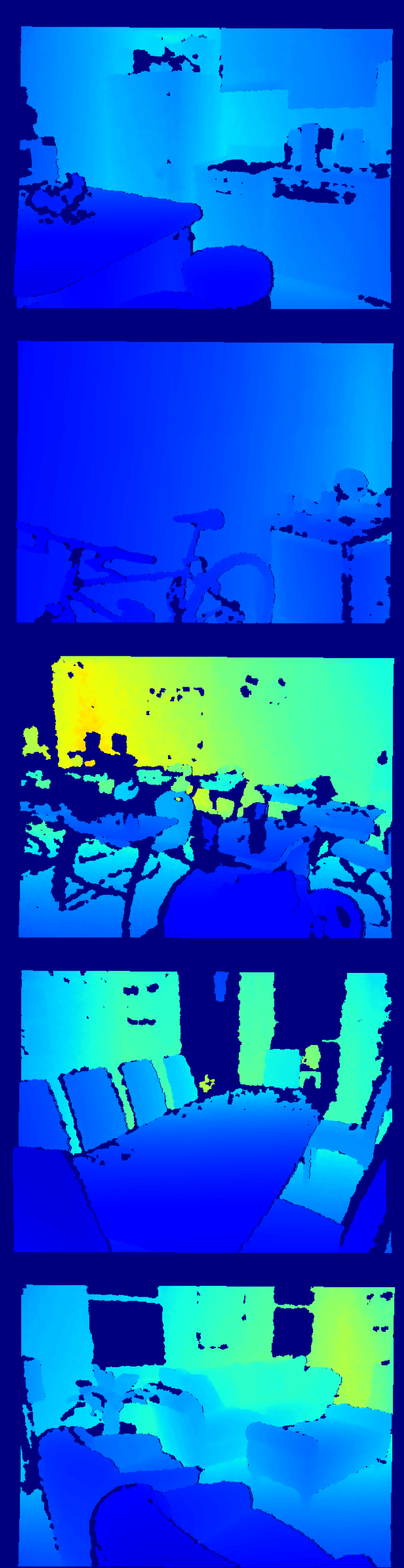}}
        \caption{Qualitative comparison with the state-of-the-art on the NYU-Depth-v2 dataset.}
        \label{fig:qualitative-comparison}
\end{figure*}

\begin{figure*}
     \centering
     \subcaptionbox{RGB}{\includegraphics[height=5cm]{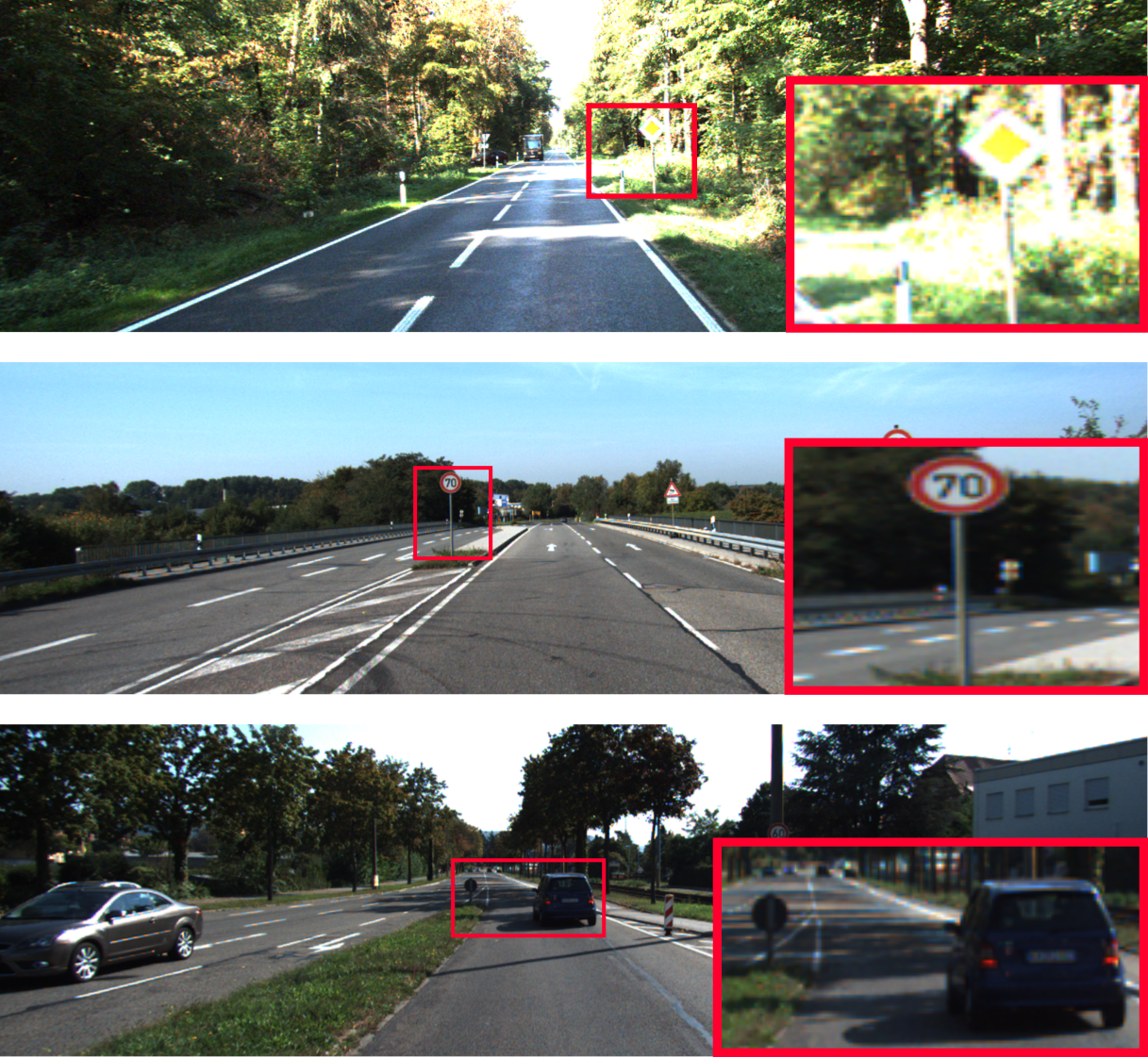}}\hspace{0.3em}%
     \subcaptionbox{BTS~\cite{bts_lee2019big}}{\includegraphics[height=5cm]{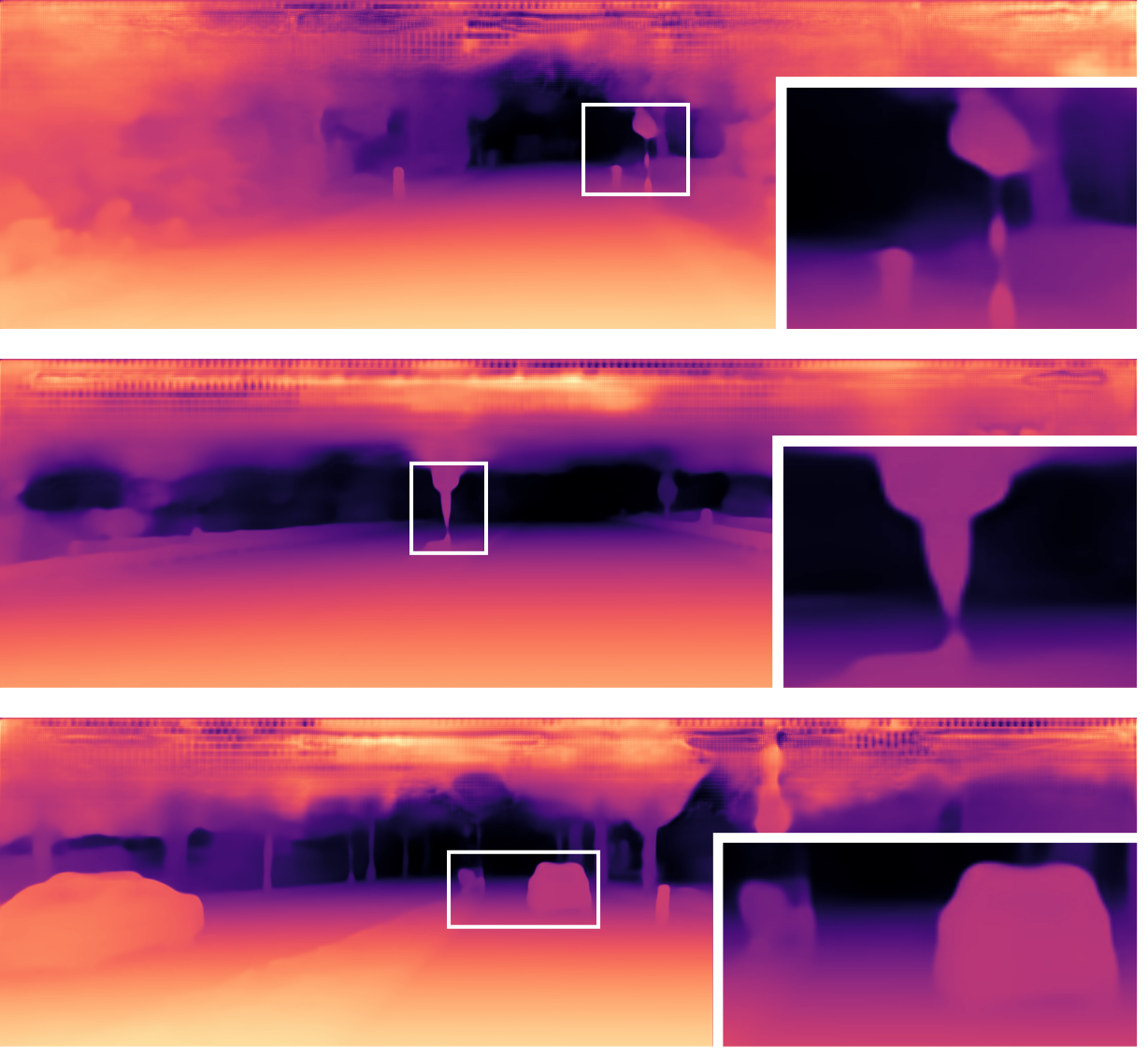}}\hspace{0.3em}%
     \subcaptionbox{Ours}{\includegraphics[height=5cm]{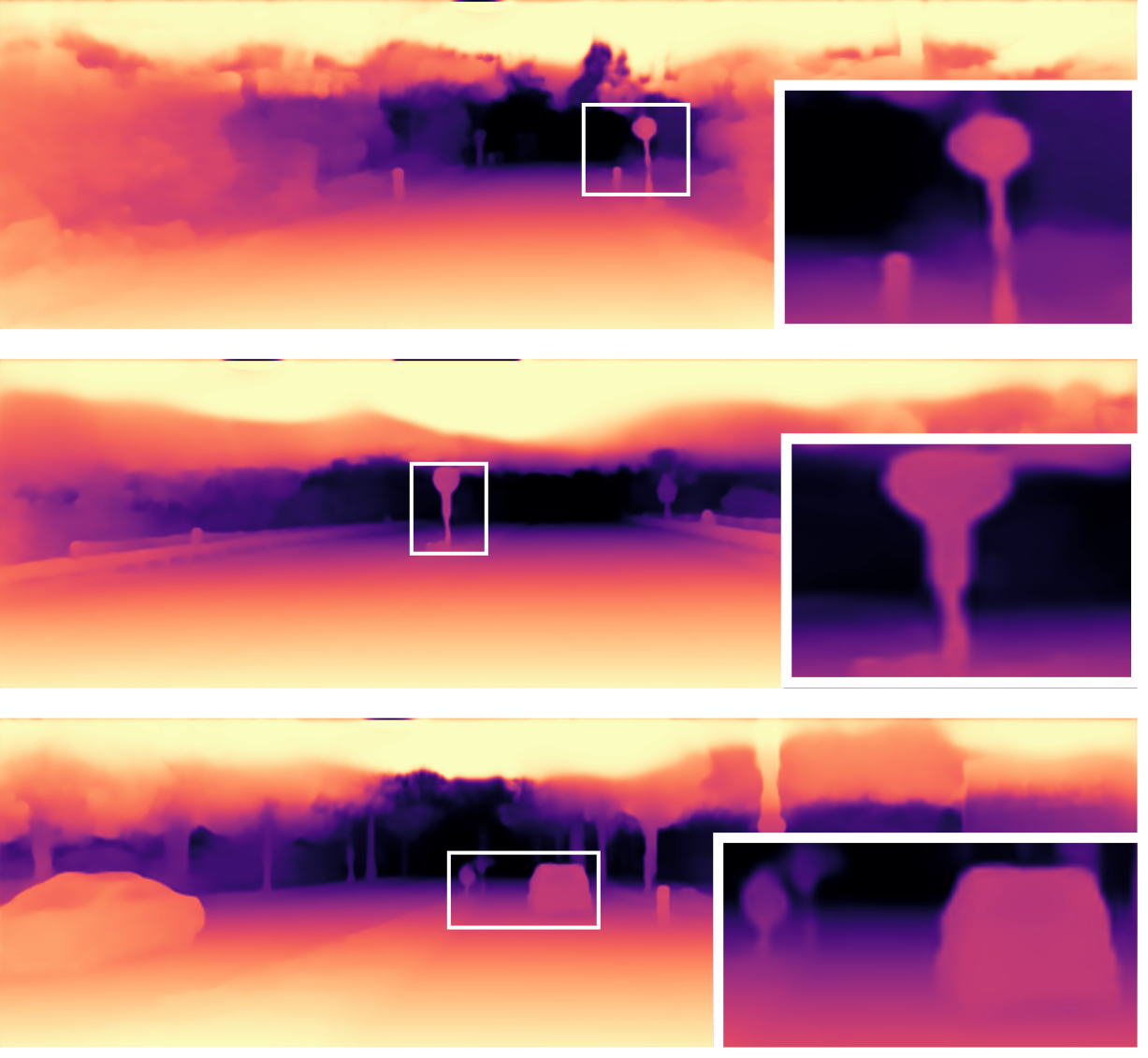}}\hspace{0.3em}%
        \caption{Qualitative comparison with the state-of-the-art on the KITTI dataset.}
        \label{fig:qualitative-comparison-kitti}
\end{figure*}

We consider the following two methods to be our main competitors: BTS~\cite{bts_lee2019big} and DAV~\cite{dav_huynh2020guiding}.  
For completeness, we also include selected previous related methods in the comparison tables.
For BTS and DAV we report the corresponding evaluation numbers from their papers. For BTS we also verified these numbers by retraining their network using the authors code. DAV did not have code available by the deadline, but the authors sent us the resulting depth images used in our figures. In our tables we report the numbers given by the authors in their paper \footnote{The authors of DAV clarified in an email that they compute the depth maps at 1/4th the resolution and then downsample the ground truth for evaluation. However, we believe that all other methods, including ours, evaluate at the full resolution.}.

\textbf{NYU-Depth-v2:} 
See Table~\ref{tab:results-nyu} for the comparison of the performance on the official NYU-Depth-v2 test set. While the state of the art performance on NYU has been saturated for quite some time, we were able to significantly outperform the state of the art in all metrics. The large gap to the previous state of the art emphasises that our proposed architecture addition makes an important contribution to improving the results.

\textbf{KITTI:} 
Table~\ref{tab:results-kitti} lists the performance metrics on the KITTI dataset. Our proposed architecture significantly outperforms previous state-of-the-art across all metrics. In particular, our method improves the RMS score by about $13.5\%$ and Squared Relative Difference by $22.4\%$ over the previous state-of-the-art.

\textbf{SUN RGB-D:}
To compare the generalisation performance, we perform a cross-dataset evaluation by training our network on the NYU-Depth-v2 dataset and evaluate it on the test set of the SUN RGB-D dataset without any fine-tuning. For comparison, we also used the same strategy for competing methods for which pretrained models are available~\cite{bts_lee2019big,Yin_2019_ICCV,ijcai2019-98} and report results in Table.~\ref{tab:generalization}.

\begin{table}[t]
\centering
\begin{adjustbox}{width=\linewidth,center}
\begin{tabular}{@{}lllllll@{}}
\toprule
Method        & $\delta_1\uparrow$     & $\delta_2\uparrow$             & $\delta_3\uparrow$          & REL$\downarrow$         & RMS$\downarrow$  & $log_{10}\downarrow$\\ \midrule
Chen~\cite{ijcai2019-98}        & \underline{0.757}          & \underline{0.943}          & \textbf{0.984}         & \underline{0.166}          & \underline{0.494}  &  \underline{0.071}\\
Yin~\cite{Yin_2019_ICCV}        & 0.696          & 0.912          & 0.973          & 0.183          &  0.541  &  0.082\\
BTS~\cite{bts_lee2019big}            & 0.740          & 0.933          & 0.980          & 0.172          & 0.515  &  0.075 \\ \midrule
\textbf{Ours}                       & \textbf{0.771}  & \textbf{0.944} & \underline{0.983} & \textbf{0.159} & \textbf{0.476}  &  \textbf{0.068} \\
\bottomrule
\end{tabular}
\end{adjustbox}
\caption{Results of models trained on the NYU-Depth-v2 dataset and tested on the SUN RGB-D dataset \cite{Song2015_sunrgbd} without fine-tuning.}
\label{tab:generalization}
\end{table}

\begin{table}[t]
\centering
\begin{adjustbox}{width=\linewidth}
\begin{tabular}{@{}llllll@{}}
\toprule
\textbf{Variant}        & \textbf{$\delta_1$}$\uparrow$       & \textbf{$\delta_2$}$\uparrow$          & \textbf{$\delta_3$}~$\uparrow$            & REL~$\downarrow$          & RMS~$\downarrow$ \\ \midrule
Base + R          & 0.881          & 0.980          & 0.996          & 0.111            & 0.419                 \\
Base + Uniform-Fix-HR                         & 0.892          & 0.981          & 0.995          & 0.107            & 0.383                \\
\begin{tabular}[c]{@{}l@{}}Base + Log-Fix-HR\end{tabular} & 0.896          & 0.981          & 0.995          & 0.108            & 0.379                  \\
\begin{tabular}[c]{@{}l@{}}Base + Train-Fix-HR\end{tabular}     & 0.893          & 0.981          & 0.995          & 0.109            & 0.381                \\
\textbf{Base + AdaBins-HR}                                                                 & \textbf{0.903} & \textbf{0.984} & \textbf{0.997} & \textbf{0.103}   & \textbf{0.364} \\
\bottomrule
\end{tabular}
\end{adjustbox}
\caption{Comparison of different design choices for bin-widths and regression. AdaBins module results in a significant boost in performance. Base: encoder-decoder with an EfficientNet B5 encoder. R: standard regression. HR: Hybrid Regression. (Log)Uniform-Fix: Fixed (log) uniform bin-widths. Train-Fix: Trained bin-widths but Fixed for each dataset.}
\label{tab:ablation-arch}
\end{table}

\subsection{Ablation study}

For our ablation study, we evaluate the influence of the following design choices on our results:\par

\textbf{AdaBins}: We first evaluate the importance of our AdaBins module. We remove the AdaBins block from the architecture and use the encoder-decoder to directly predict the depth map by setting $C_d=1$. We then use the loss given by Eq.~\ref{eq:pixel-loss} to train the network. We call this design \textit{standard regression} and compare it against variants of our AdaBins module. Table.~\ref{tab:ablation-arch} shows that the architecture without AdaBins (Row 1) performs worse than all other variants (Rows 2-5). \par

\textbf{Bin types}:
In this set of experiments we examine the performance of adaptive bins over other choices as stated in Sec.~\ref{sec:adabins-design-choices}. Table.~\ref{tab:ablation-arch} lists results for all the discussed variants. The Trained-but-Fixed variant performs worst among all choices and our final choice employing adaptive bins significantly improves the performance and outperforms all other variants.\par

\textbf{Number of bins ($N$)}: To study the influence of the number of bins, we train our network for various values of $N$ and measure the performance in terms of Absolute Relative Error metric. Results are plotted in Fig.~\ref{fig:ablation-N}. Interestingly, starting from $N=20$, the error first increases with increasing $N$ and then decreases significantly. As we keep increasing $N$ above $256$, and with higher values the gain in performance starts to diminish. We use $N=256$ for our final model.\par

\textbf{Loss function}: Table.~\ref{tab:ablation-loss} lists performance corresponding to the three choices of loss function. Firstly, the $L_1$/SSIM combination does not lead to the state-of-the-art performance in our case. Secondly, we trained our network with and without the proposed Chamfer loss (Eq.~\ref{eq:chamfer-loss}). Introducing the Chamfer loss clearly gives a boost to the performance. For example, introducing the Chamfer loss reduces the Absolute Relative Error from $10.6\%$ to $10.3\%$.

\section{Conclusion}
We introduced a new architecture block, called \emph{AdaBins} for depth estimation from a single RGB image. AdaBins leads to a decisive improvement in the state of the art for the two most popular datasets, NYU and KITTI.
In future work, we would like to investigate if global processing of information at a high resolution can also improve performance on other tasks, such as segmentation, normal estimation, and 3D reconstruction from multiple images.



{\small
\bibliographystyle{ieee_fullname}
\bibliography{egbib}
}

\appendix

\section{Appendix}
\subsection{Geometric Consistency}
\begin{figure*}[h]
    \centering
    \begin{tabular}{l}
         \includegraphics[width=\linewidth]{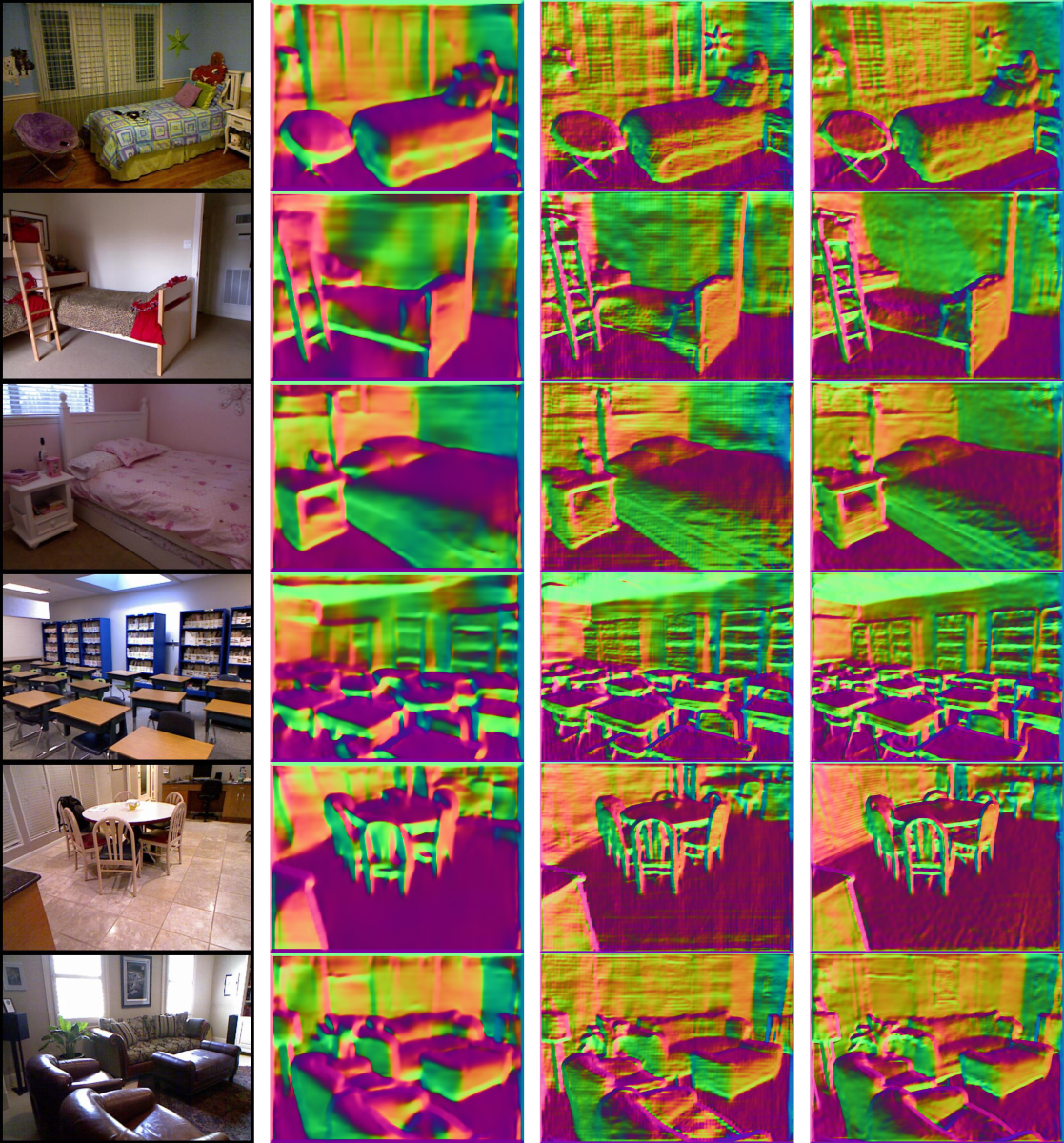} \\
         \hspace{0.1\linewidth}RGB \hspace{0.2\linewidth}DAV~\cite{dav_huynh2020guiding}\hspace{0.18\linewidth}BTS~\cite{bts_lee2019big} \hspace{0.19\linewidth}Ours\\
    \end{tabular}
    
    \caption{Visualization of surface normals extracted from predicted depth maps.}
    \label{fig:sup_normals}
\end{figure*}

We provide a qualitative evaluation of the geometric consistency of depth maps predicted by our model. Surface normal maps provide a good way to visualize the orientation and texture details of surfaces present in the scene. Fig~\ref{fig:sup_normals} shows the visualization of the normals extracted from the depth maps for our model and for DAV~\cite{dav_huynh2020guiding} and BTS~\cite{bts_lee2019big}. Although the orientations predicted by DAV seems to be consistent, the texture details are almost completely lost. BTS, on the other hand, preserves the texture but sometimes results in erroneous orientation details. Our method exhibits detailed texture and consistent orientations without explicitly imposing geometric constraints, such as co-planarity, used by other  methods~\cite{dav_huynh2020guiding, bts_lee2019big}.

\subsection{Generlization Analysis}
\begin{figure*}[t]
    \centering
    \begin{tabular}{c}
         \includegraphics[width=0.85\linewidth]{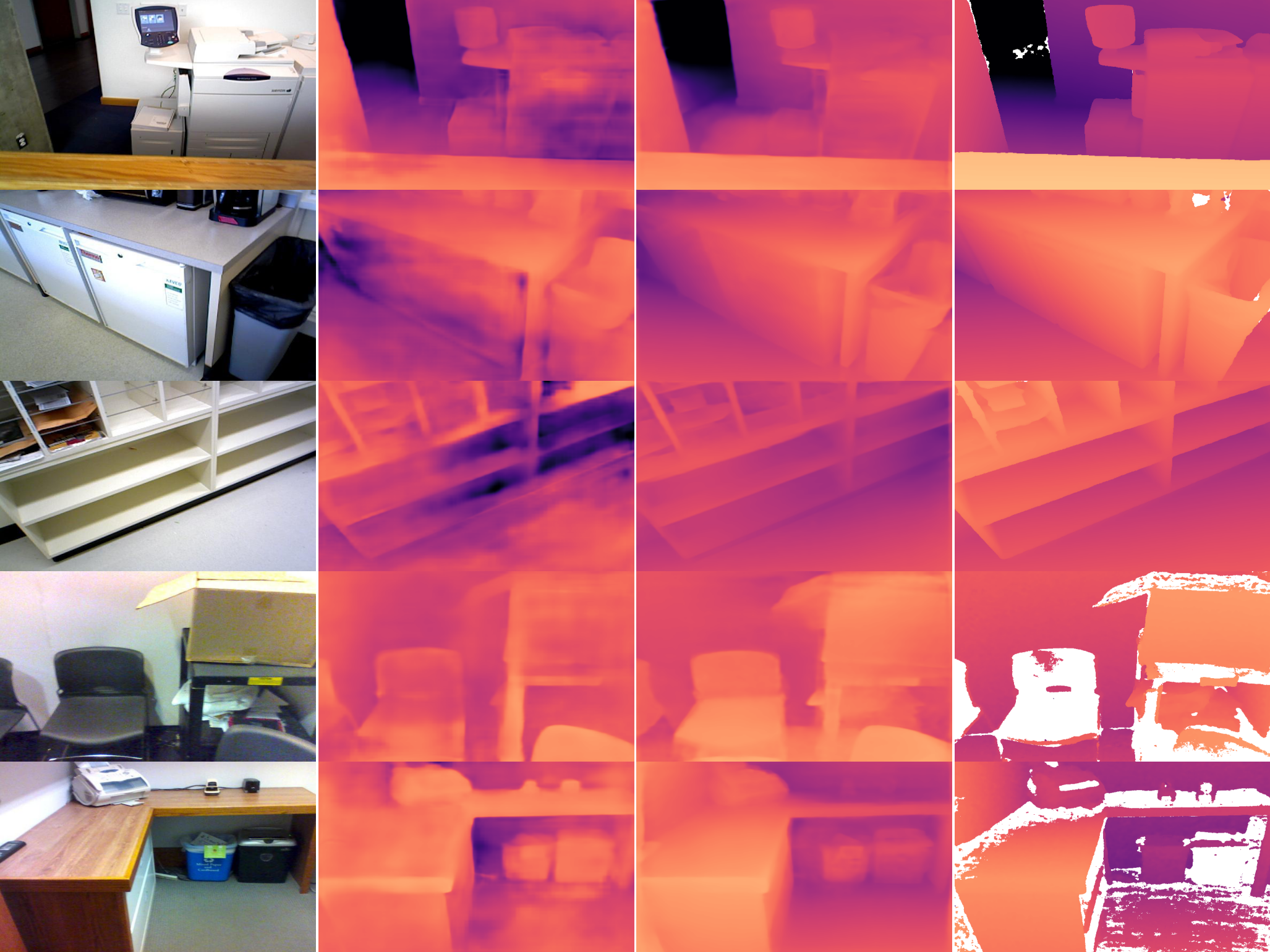} \\
         \hspace{0.0\linewidth}RGB \hspace{0.15\linewidth}BTS~\cite{bts_lee2019big}\hspace{0.14\linewidth}Ours \hspace{0.17\linewidth}GT\\
    \end{tabular}
    
    \caption{Qualitative comparison of generalization from NYU-Depth-v2 to SUN RGB-D dataset. 
    Darker pixels are farther. Missing ground truth values are shown in white.}
    \label{fig:sup_sun_rgbd_comparison}
\end{figure*}

Here we qualitatively analyze the capability of our method to generalise to unseen data. We use the models (AdaBins and BTS~\cite{bts_lee2019big}) trained on NYU-Depth-v2~\cite{Silberman2012} but show predictions on SUN RGB-D~\cite{Song2015_sunrgbd} dataset in Fig~\ref{fig:sup_sun_rgbd_comparison}. Depth maps predicted by BTS have conspicuous artifacts whereas our method provides consistent results on the unseen data.

\subsection{More Results on KITTI dataset}
\begin{figure*}
    \centering
    \begin{tabular}{l}
         \includegraphics[width=\linewidth]{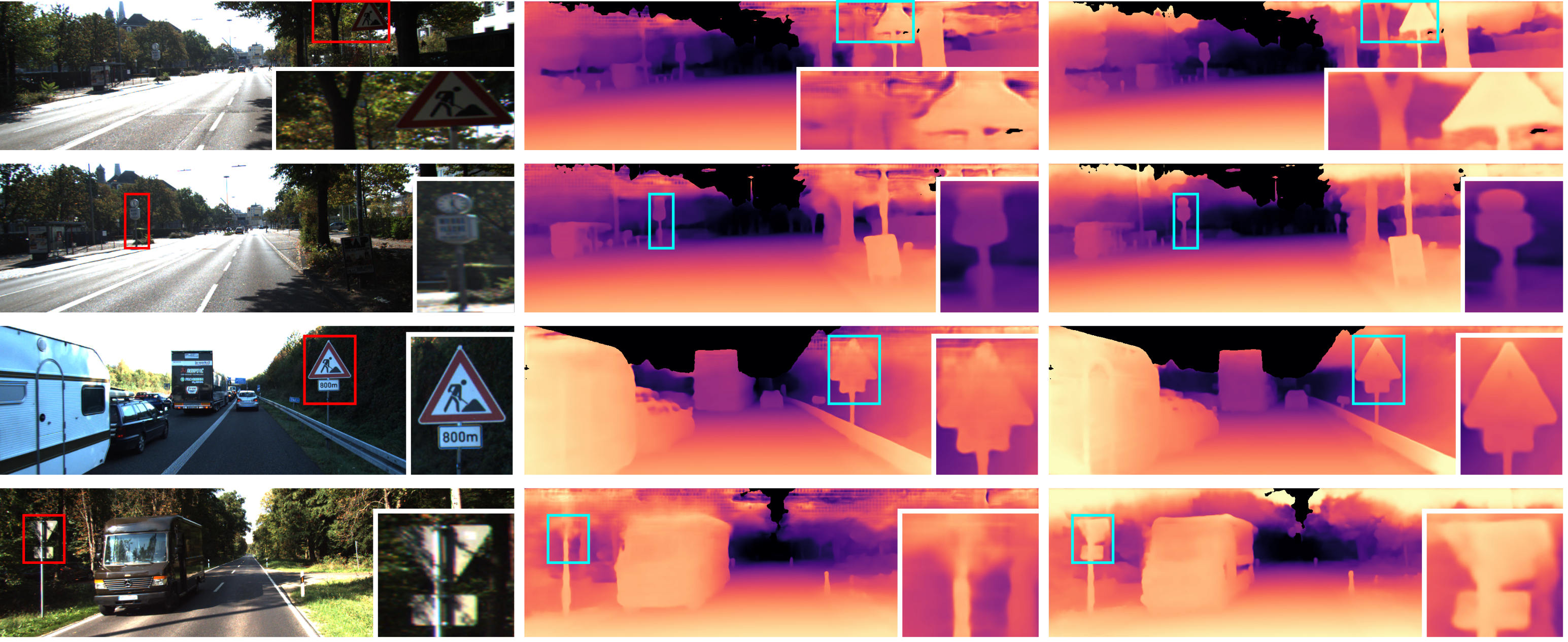} \\
         \hspace{0.13\linewidth}RGB \hspace{0.28\linewidth}BTS~\cite{bts_lee2019big} \hspace{0.28\linewidth}Ours\\
    \end{tabular}
    
    \caption{Qualitative comparison on KITTI dataset.}
    \label{fig:sup_kitti_comparison}
\end{figure*}

Fig~\ref{fig:sup_kitti_comparison} shows a qualitative comparison of BTS~\cite{bts_lee2019big} and our method on the KITTI dataset. For better visualization, we have removed the sky regions from the visualized depth maps using segmentation masks predicted by a pretrained segmentation model\cite{deeplabv3plus2018}. We can observe that our method demonstrates superior performance particularly in predicting extents and edges of the on-road vehicles, sign-boards and thin poles. Additionally, BTS tends to blend the farther away objects with background whereas our method preserves the structure with clear separation.

\subsection{MLP Head Details}
We use a three-layer MLP on the first output embedding of the transformer in the mini-ViT module. The architecture details with parameters are given in Table~\ref{tab:sup_arch_mlp_head}.

\pagebreak
\begin{table}[h]
\centering
\begin{adjustbox}{width=\linewidth}

\begin{tabular}{@{}lccl@{}}
\toprule
Layer & Input Dimension & Output Dimension & Activation                                                                 \\ \midrule
FC    & E               & 256              & \begin{tabular}[c]{@{}l@{}}LeakyReLU\\ \small(negative\_slope=0.01)\end{tabular} \\ 
FC    & 256             & 256              & \begin{tabular}[c]{@{}l@{}}LeakyReLU\\  \small(negative\_slope=0.01)\end{tabular} \\ 
FC    & 256             & N                & ReLU  \\
\bottomrule
\end{tabular}
\end{adjustbox}
\caption{Architecture details of MLP head. FC: Fully Connected layer, E: Embedding dimension, N: Number of bins}
\label{tab:sup_arch_mlp_head}
\end{table}



\end{document}